\documentclass{styles/svproc}
\usepackage{url}

\usepackage{graphicx}
\usepackage{epsfig}
\usepackage{amsmath}
\usepackage{amssymb}
\usepackage{enumitem}
\usepackage{booktabs}
\usepackage{multirow}

\usepackage{bm}
\usepackage{upgreek}
\usepackage{algorithmic}
\usepackage[vlined,ruled]{algorithm2e}
\usepackage{soul,xcolor}
\usepackage[bottom]{footmisc}
\usepackage{mathtools}
\usepackage[breaklinks=true,colorlinks=true,citecolor=blue]{hyperref}
\spnewtheorem{rmk}{Remark}[section]{}{\itshape}

\DeclareMathOperator*{\avg}{avg}

\allowdisplaybreaks

\begin{document}
\mainmatter              % start of a contribution
\title{Bayesian Sampling Bias Correction: \\Training with the Right Loss Function}
\titlerunning{Bayesian Sampling Bias Correction}  % abbreviated title (for running head)

\author{Lo\"ic Le Folgoc\inst{1} 
\and
Vasileios Baltatzis\inst{2} \and
Amir Alansary\inst{1} \and
Sujal Desai\inst{1,3} \and
Anand Devaraj\inst{3} \and
Sam Ellis\inst{3} \and
Octavio E. Martinez Manzanera\inst{2} \and
Fahdi Kanavati\inst{1} \and
Arjun Nair\inst{4} \and	
Julia Schnabel\inst{2} \and
Ben Glocker\inst{1}}
\authorrunning{L. Le Folgoc et al.}
\institute{
BioMedIA, Imperial College London, United Kingdom \\
\and Biomedical Engineering and Imaging Sciences, King’s College London, UK \\
\and The Royal Brompton \& Harefield NHS Foundation Trust, London UK \\
\and Department of Radiology, University College London, UK \\
\email{l.le-folgoc@imperial.ac.uk}
}

\tocauthor{}
%Craig Chambers, Kim B. Bruce, and Elisa Bertino}
%

\maketitle              % typeset the title of the contribution

% Added macros
%!TEX root = paper.tex

\newcommand{\R}{\mathbb{R}}
\newcommand{\bbC}{\mathbb{C}}
\newcommand{\bbH}{\mathbb{H}}
\newcommand{\bbi}{\boldsymbol{\mathrm{i}}}
\newcommand{\bbj}{\boldsymbol{\mathrm{j}}}
\newcommand{\bbk}{\boldsymbol{\mathrm{k}}}
\newcommand{\calB}{\mathcal{B}}
\newcommand{\calC}{\mathcal{C}}
\newcommand{\calD}{\mathcal{D}}
\newcommand{\calE}{\mathcal{E}}
\newcommand{\calF}{\mathcal{F}}
\newcommand{\calG}{\mathcal{G}}
\newcommand{\calH}{\mathcal{H}}
\newcommand{\calI}{\mathcal{I}}
\newcommand{\calL}{\mathcal{L}}
\newcommand{\calM}{\mathcal{M}}
\newcommand{\calN}{\mathcal{N}}
\newcommand{\calO}{\mathcal{O}}
\newcommand{\calQ}{\mathcal{Q}}
\newcommand{\calR}{\mathcal{R}}
\newcommand{\calS}{\mathcal{S}}
\newcommand{\calT}{\mathcal{T}}
\newcommand{\calV}{\mathcal{V}}
\newcommand{\calX}{\mathcal{X}}
\newcommand{\calY}{\mathcal{Y}}
\newcommand{\calW}{\mathcal{W}}
\newcommand{\rmA}{\mathrm{A}}
\newcommand{\rmC}{\mathrm{C}}
\newcommand{\rmD}{\mathrm{D}}
\newcommand{\rmE}{\mathrm{E}}
\newcommand{\rmF}{\mathrm{F}}
\newcommand{\rmH}{\mathrm{H}}
\newcommand{\rmI}{\mathrm{I}}
\newcommand{\rmJ}{\mathrm{J}}
\newcommand{\rmL}{\mathrm{L}}
\newcommand{\rmM}{\mathrm{M}}
\newcommand{\rmQ}{\mathrm{Q}}
\newcommand{\rmR}{\mathrm{R}}
\newcommand{\rmU}{\mathrm{U}}
\newcommand{\rme}{\mathrm{e}}
\newcommand{\rmf}{\mathrm{f}}
\newcommand{\rmh}{\mathrm{h}}
\newcommand{\rmn}{\mathrm{n}}
\newcommand{\rmp}{\mathrm{p}}
\newcommand{\rmq}{\mathrm{q}}
\newcommand{\rmu}{\mathrm{u}}
\newcommand{\rmv}{\mathrm{v}}
\newcommand{\rmw}{\mathrm{w}}
\newcommand{\rmx}{\mathrm{x}}
\newcommand{\rmy}{\mathrm{y}}
\newcommand{\bmc}{\bm{c}}
\newcommand{\bme}{\bm{e}}
\newcommand{\bmf}{\bm{f}}
\newcommand{\bmh}{\bm{h}}
\newcommand{\bms}{\bm{s}}
\newcommand{\bmq}{\bm{q}}
\newcommand{\bmt}{\bm{t}}
\newcommand{\bmr}{\bm{r}}
\newcommand{\bmu}{\bm{u}}
\newcommand{\bmw}{\bm{w}}
\newcommand{\bmy}{\bm{y}}
\newcommand{\bmz}{\bm{z}}
\newcommand{\brmf}{\boldsymbol{\rmf}}
\newcommand{\brmh}{\boldsymbol{\rmh}}
\newcommand{\brmp}{\boldsymbol{\rmp}}
\newcommand{\brmx}{\boldsymbol{\rmx}}
\newcommand{\brmy}{\boldsymbol{\rmy}}
\newcommand{\brmA}{\boldsymbol{\rmA}}
\newcommand{\brmC}{\boldsymbol{\rmC}}
\newcommand{\brmD}{\boldsymbol{\rmD}}
\newcommand{\brmE}{\boldsymbol{\rmE}}
\newcommand{\brmH}{\boldsymbol{\rmH}}
\newcommand{\brmI}{\boldsymbol{\rmI}}
\newcommand{\brmL}{\boldsymbol{\rmL}}
\newcommand{\brmQ}{\boldsymbol{\rmQ}}
\newcommand{\brmR}{\boldsymbol{\rmR}}
\newcommand{\dis}{\displaystyle}
\newcommand{\T}{{\mkern-1.5mu\mathsf{T}}}
\newcommand{\Id}{\text{Id}}
\newcommand{\tr}{\text{tr}}
\newcommand{\etal}{{\it et al }}
\newcommand{\eg}{\textit{e.g.}\xspace}
\newcommand{\ie}{\textit{i.e.}\xspace}
\newcommand{\uline}[1]{\underline{#1}}
\newcommand{\duline}[1]{\underline{\underline{#1}}}
\newcommand{\eq}{\!=\!}
\newcommand{\mrho}{{\!\rho}}
\newcommand{\iid}{{\textit{i.i.d.}}}

%%%%%%%%%%%%%%%%%% paper specific

\def\tool{{\textsc{Eugene}}}

\begin{abstract}
We derive a family of loss functions to train models in the presence of sampling bias.  
Examples are when the prevalence of a pathology differs from its sampling rate in the training dataset, or when a machine learning practioner rebalances their training dataset.
Sampling bias causes large discrepancies between model performance in the lab and in more realistic settings. 
It is omnipresent in medical imaging applications, yet is often overlooked at training time or addressed on an ad-hoc basis. 
Our approach is based on Bayesian risk minimization. 
For arbitrary likelihood models we derive the associated bias corrected loss for training, exhibiting a direct connection to information gain. 
The approach integrates seamlessly in the current paradigm of (deep) learning using stochastic backpropagation and naturally with Bayesian models. 
We illustrate the methodology on case studies of lung nodule malignancy grading.

\end{abstract}

\section{Introduction}

Much of MICCAI literature consists of observational, case-control studies. Given training data $X,Y\!\in\! \calX\!\times\!\calY$ one 
learns to predict a dependent variable $y\!\in\!\calY$ (\eg the classification label) from inputs $x\!\in\!\calX$ (a.k.a., \textit{covariates} or features), optimally for the population distribution $x_\ast,y_\ast\!\sim\!\calD$. 
Key to our work, the predictive power of the predictor depends on the marginal statistics of the population of interest. For instance the precision of a test depends on the prevalence of the disease; the accuracy of a classifier depends on the class (im)balance. 
Sampling bias is the discrepancy between the distribution $\calD'$ of the training dataset and the distribution $\calD$ of the actual population of interest.
It affects the accuracy of predictive inference (\eg classification accuracy on $\calD$) and of statistical findings (\eg the strength of association between exposure and outcome).
The training set $\calD'$ results from a complex process. There are numerous sampling protocols for data collection (\eg random, stratified, clustered, subjective) and the machine learning (ML) practitioner may further adjust the dataset at training time.

Sampling bias can be introduced at either stage. ML practice tends to repurpose retrospective data in a way that mismatches the original study, or unaware of inclusion/exclusion criteria specific to that study. An automated screening model may be trained from incidental data or from purpose-made data collected in specialized units. Incidence rates would differ between these populations and the general population. Statistics may further be biased by the acquisition site \eg, by country, hospital; and by practical choices. Say, clinical partners may handcraft a balanced dataset with equal amounts of healthy and pathological cases; relying on their expertise to judge the value and usefulness of a sample \eg, discarding trivial or ambiguous cases (\textit{subjective} sampling), or based on quality control criteria (\eg, image quality). 
At the other end the ML practitioner chasing quantifiable performance on their dataset is also likely to design sampling heuristics that disregard true population statistics. Such performance gains may not transfer to the real world.

\begin{figure}[t]
\includegraphics[width=\textwidth]{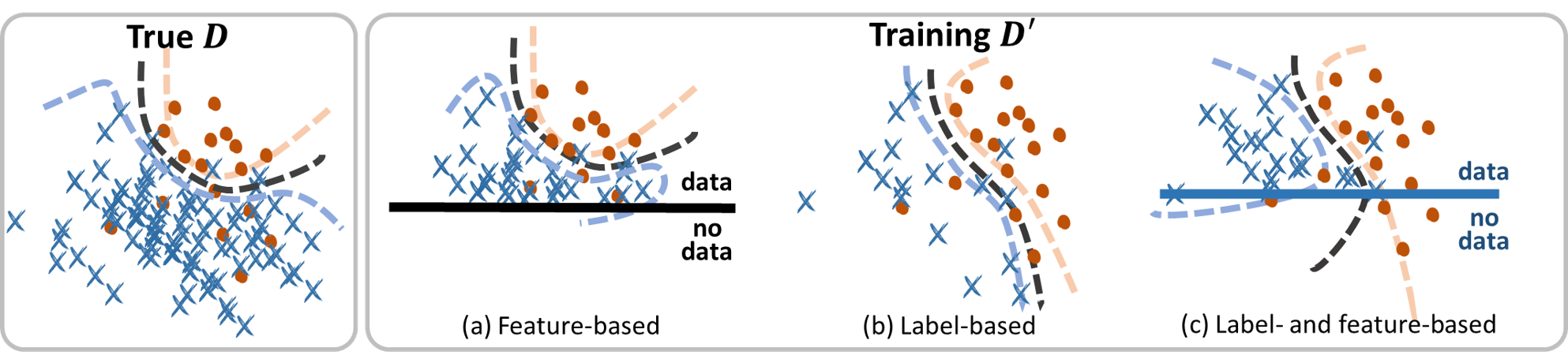}
\caption{Sampling bias occurs when the true distribution $\calD$ of the population of interest $x_\ast,y_\ast\!\sim\!\calD$ differs from the training distribution $\calD'$ of the dataset $X,Y$. (a) Feature-based sampling \eg, no data collected for children and elderly people; or subjects for which the diagnostic is trivial not included. (b) Label-based sampling \eg, size of the dataset fixed to $1000$ samples, evenly split between benign and malignant; dataset from oncology unit used for general screening tasks. (c) General setting \eg, patient data ($y\eq1$) collected for all groups upon visiting clinics; volunteer data ($y\eq 0$) restricted to groups with incentive to join. (Worst case, sampling bias unknown.)} \label{fig: sampling biases}
\end{figure}

Heckman~\cite{heckman1979sample} provides in Nobel Prize winning econometrics work a comprehensive discussion of, and methods for analyzing selective samples. The typology is adopted in sociology~\cite{berk1983introduction}, machine learning~\cite{zadrozny2004learning,cortes2008sample} and for statistical tests in genomics~\cite{young2010gene} and medical communities~\cite{stukel2007analysis}. Selection biases are discussed from the broader scope of structural biases in sociology~\cite{winship1999estimation} and epidemiology~\cite{hernan2004structural}. In the worst case the mechanism underlying the bias is unknown; and potentially conditions both on causal variables $x$ and outcome variables $y$. Early work in this setting is for bias correction in linear regression models with fully parametric or semi-nonparametric selection models~\cite{vella1998estimating}. We focus instead on practical scenarii with some knowledge of the bias but arbitrary nonlinear relashionships between covariates $x$ and the dependent variable $y$ (Fig.~\ref{fig: sampling biases}). Section~\ref{sec: method} formalizes the precise setting. 

The paper focuses on the case of \textit{label}-based sampling as in~\cite{elkan2001foundations,lin2002support} and Fig.~\ref{fig: sampling biases}(b), but unlike~\cite{shimodaira2000improving,zadrozny2004learning,huang2007correcting,cortes2014domain} who address \textit{covariate} shift (Fig.~\ref{fig: sampling biases}(a)). 
Our approach is derived from Bayesian principles. 
From this standpoint undersampling parts of the input space $\calX$ would mostly result in higher uncertainty, whereas undersampling a specific label invalidates the (probabilistic) decision boundary. 
The phenomenon is well-known and motivates the use of sensitivity-specificity plots (ROC curves) after training to determine the best operating point. 
But can label-based selection bias (mismatched prevalence in $\calD$ and $\calD'$) be accounted for at training time?
Much of the machine learning literature~\cite{shimodaira2000improving,elkan2001foundations,lin2002support,zadrozny2004learning,huang2007correcting} adopts a strategy of importance weighting, whereby the cost of training sample errors is weighted to more
closely reflect that of the test distribution.
Importance weighting is rooted in regularized risk minimization~\cite{huang2007correcting}, that is maximizing the expected log-likelihood $\mathbb{E}_\calD[\log{p(y|x,w)}]$ plus a regularizer $-\lambda\calR(w)\!\equiv\!\log{p(w)}$, w.r.t. model parameters $w$. Our analysis departs from importance weighting. It leads instead to a modified training likelihood.
Related work also appears in the literature on transfer learning~\cite{pan2009survey,weiss2016survey,shin2016deep} and domain adaptation~\cite{cortes2014domain,kamnitsas2017unsupervised,frid2018gan} driven by NLP, speech and image processing applications. The aim is to cope with generally ill-posed shifts of the distribution of the input $x$. In that sense the present paper is orthogonal to, and can be combined with this body of work. Finally the problem of class imbalance is central in medical image segmentation where a class (\eg, the background) is often over-represented in the dataset. It brings about a number of resampling (class rebalancing) strategies, see for instance a discussion of their effect on various metrics in~\cite{kamnitsas2017efficient}, as well as a review, benchmark and informative look into various empirical corrections in~\cite{li2019overfitting}.

\section{Method}
\label{sec: method}

\begin{figure}[t]
\centering
\includegraphics[width=.75\textwidth]{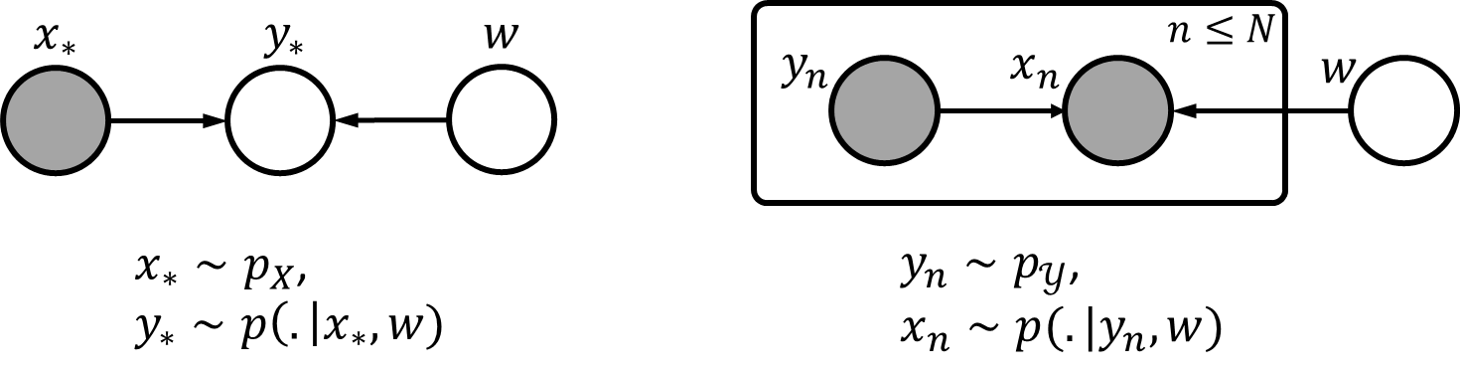}
\caption{Generative model of the true population $(x_\ast,y_\ast)\!\sim\!\calD$ (a) and of the training dataset $(x_n,y_n)\!\sim\!\calD'$ (b) under the label-based sampling bias of Fig.~\ref{fig: sampling biases}(b).}
\label{fig: graphical models}
\end{figure}

We consider the \textit{label}-based sampling bias of Fig.~\ref{fig: sampling biases}(b). The proposed approach is formalized from a Bayesian standpoint. We specify a generative model for the population of interest and for the training dataset as illustrated in Fig.~\ref{fig: graphical models}. Using these models and Bayesian risk minimization, let us anticipate the main result that given a training set $X,Y$ with label-based sampling bias and a bias-free likelihood $p(y|x,w)$, the posterior on model parameters $w$ is expressed as:
\begin{equation}
{p(w|X,Y)} \enspace \propto \enspace 
\underbrace{%
\vphantom{\frac{p(y_n|x_n,w)}{p(y_n|w)} }
p(w)
}_{\text{prior}} 
\; \cdot \; 
\prod_{n\leq N} \!\!\underbrace{%
\frac{p(y_n|x_n,w)}{p(y_n|w)}
}_{\substack{\text{surrogate training}\\ \text{likelihood}}}\, ,
\label{eq: bias-corrected posterior}
\end{equation}
indexing samples by $n$. The bias-corrected posterior includes a normalizing factor in the denominator of the surrogate likelihood, the marginal $p(y_n|w)$\footnote{Namely $p(y_n|w)\eq\int_\calX p(y_n|x'_n,w)p_\calX(x'_n)dx'_n$}. Hence it captures the relative \textit{information gain} when conditioning on $x_n$ compared to a ``random guess'' based on marginal statistics. 
Besides $p(w|X,Y)$ can be approximated using any standard strategy from MAP to VI~\cite{blundell2015weight}, EP~\cite{minka2013expectation}, MCMC~\cite{chen2016bridging}. At test time the (approximate) posterior is combined with the likelihood $p(y_\ast|x_\ast,w)$ as usual to yield the Bayesian risk-minimizing predictive posterior $p(y_\ast|x_\ast,X,Y)$. In practical DL terms, 
one trains the NN with the surrogate loss $\calL_n^{\text{BC}}(w)\triangleq \log\{p(y_n|x_n,w)/p(y_n|w)\}$ instead of the usual loss $\calL_n(w)\!\triangleq\! \log{p(y_n|x_n,w)}$ to find the optimal weights $\hat{w}$. At test time the standard likelihood $p(y_n|x_n,\hat{w})$ is used for the prediction. 
The only practical point to address is the computation of $p(y_n|w)$, and the reader can skip directly to the relevant section.\\

\noindent
\textbf{Generative model.} For the \textbf{true population} $x_\ast,y_\ast\!\sim\!\calD_w$\footnote{$D\!\triangleq\!\calD_w$ in the informal discussion that preceeds. We now explicitly index by $w$.} of interest at test time (Fig.~\ref{fig: graphical models}(a)), the dependent variable $y_\ast\!\sim\! p(y_\ast|x_\ast,w)$ is caused by $x_\ast$, according to a probabilistic model $x\mapsto p(\cdot|x,w)$ with unknown parameters $w$\footnote{Say, for binary classification the standard model is $x\mapsto \calB(\sigma[\text{NN}_w(x)])$ \ie, the label results from a Bernoulli draw. The probability of $y\eq 1$ is obtained by squashing the output of a neural network architecture $\text{NN}_w(x)$ through a logistic link function $\sigma$.}. For instance age, sex and life habits ($x$) may condition the probability of developing cancer ($y$). Image data ($x$) might condition the patient management ($y$). The sampling process $x_\ast,y_\ast\!\sim\!\calD_w$ intuitively expands as $x_\ast\!\sim\!p_\calX$, $y_\ast\!\sim\!p(\cdot |x_\ast,w)$. The marginal distribution $p(y_\ast|w)\eq\int_\calX p(y_\ast|x_\ast,w)p_\calX(x_\ast)dx_\ast$ of $y_\ast$ depends on the population distribution $p_\calX$, but $p(y_\ast|x_\ast,w)$ does not. It remains unaffected by population drift ($\triangleq$ change of $p_\calX$).

The \textbf{training} dataset $X,Y$ follows a different generative process (Fig.~\ref{fig: graphical models}(b)) by assumption of label-based sampling. 
Labels $Y\!\sim\! \tilde{p}(Y)$ are sampled first. The notation emphasizes that the distribution $\tilde{p}$ is linked to the training dataset design and should not be confused with say, $p(y)\eq\int_w p(y|w)p(w)dw$. Then $x_n\!\sim\!p(x_n|y_n,w)$ is drawn uniformly \ie, according to the true conditional distribution $\calD_{x|y_n,w}$, since the selective bias only involves $y$.\\

\noindent
\textbf{Bayesian risk minimization.} We want an optimal prediction rule $q_{x_\ast,X,Y}(y)$ for new observations $x_\ast\!\sim\!p_\calX$ given training data $X,Y\!\sim\!\calD'_w$. A prediction rule $q_{x_\ast,X,Y}$ is a probability distribution over $\calY$ that is a function of $x_\ast,X,Y$. The ideal prediction rule would be optimal w.r.t. the prediction risk $\log{q_{x_\ast,X,Y}(y_\ast)}$ on expectation over $(x_\ast,y_\ast)\sim\!\calD_w$. Since the true model parameters $w$ are unknown, the Bayes prediction risk is the expected prediction risk w.r.t. the prior distribution $p(w)$ of $w$:
\begin{equation}
\calL_{\text{Bayes}}[q_{x_\ast,X,Y}(y)] = 
-\mathbb{E}_{w\sim p(w)}\!\left[
	\mathbb{E}_{(X,Y)\sim\calD'_w}\!\left[
		\mathbb{E}_{(x_\ast,y_\ast)\sim\calD_w}[\log{q_{x_\ast,X,Y}(y_\ast)}]
	\right]
\right] \, .
\label{eq: Bayes prediction risk}
\end{equation}  
The posterior predictive distribution $p(y_\ast|x_\ast,X,Y)$ minimizes $\calL_{\text{Bayes}}$ as usual (Appendix~\ref{sec: proofs}). It expands as a weighted sum over the space of model parameters:
\begin{equation}
p(y_\ast|x_\ast,X,Y)=\int_w \underbrace{p(y_\ast|x_\ast,w)}_{\text{likelihood}}\underbrace{p(w|X,Y)}_{\text{posterior}}dw\, .
\end{equation}
The point of departure from the bias-free setting is in the exact form of the posterior $p(w|X,Y)$. 
Label-based bias induces a 
change in the structural dependencies between model variables \eg, $Y\!\perp\!\!\!\perp\! w$. The proof of Eq.~\eqref{eq: bias-corrected posterior} reported in Appendix~\ref{sec: proofs} relies on this insight. \\

\noindent
\textbf{Backpropagation through the marginal.} Stochastic backpropagation through the logarithm of Eq.~\eqref{eq: bias-corrected posterior} requires to evaluate the first-order derivative of $\log{p(y_n|w)}$. 
The following empirical estimate based on the training data holds (Appendix~\ref{sec: proofs}): 
\begin{equation}
p(y|w)\simeq\avg_n{\left(\beta_n \cdot p(y|x_n,w)\right)}\, , \quad \beta_n \coloneqq \frac{p_\calY(y_n)}{\tilde{p}(y_n)}\, , 
\label{eq: empirical marginal estimate}
\end{equation}
where $p_\calY$ is the true population marginal and $\tilde{p}(y)$ the probability of label $y$ in the training dataset. $\tilde{p}(y)$ is chosen by the user when rebalancing the training set. In absence of rebalancing it can be set to the empirical frequency of labels in $Y$. 
We assume the true population marginal to be known (\eg prevalence of a disease in the population at the time of aquiring the data). 
We derive the analytical gradient $\nabla_w \log{p(y|w)}$ of the log marginal probability: 
\begin{equation}
\nabla_w \log{p(y|w)}\simeq\avg_n{\left(\beta_n \cdot \frac{p(y|x_n,w)}{p(y|w)}\cdot\nabla_w \log{p(y|x_n,w)}\right)} \, , 
\label{eq: log marginal gradient}
\end{equation}
Eq.~\eqref{eq: log marginal gradient} is key to our implementation of the training loss. Since it is a sum of sample contributions it is easily turned into an unbiased minibatch estimate, so that gradients can flow from the loss, through the marginal, to the log-likelihood of the minibatch samples and back onto $w$. The only prerequisite is an approximation of the marginal ${p(y|w)}$ that still appears on the RHS of Eq.~\eqref{eq: log marginal gradient}.\\

\noindent
\textbf{Computation of the marginal.} The categorical distribution $p(y|w)$ is approximated by an auxiliary network $q_\psi(w)$ that takes input $w$ and returns the marginal (log-)probabilities for the $K$ labels\footnote{For regression a (parametric) density can be computed.}. We use the approximated $q_\psi(w)$  in place of $p(y|w)$ wherever it appears. 
As per Eq.~\eqref{eq: log marginal gradient} $q_{\psi}(w)$ only needs to provide an accurate $0$th-order approximation  
since its derivative is not used for backpropagation. 
The network $q_\psi$ is trained jointly with the main model, by stochastic descent over the risk $\calL_w(\psi)\!\triangleq\!\mathbb{E}_{p(\cdot|w)}[\log{q_\psi(w)}]$, equivalently $\text{KL}[p(\cdot|w)\Vert q_\psi(w)]$, on expectation over the current estimate $q(w)$ of the posterior. 
From Eq.~\eqref{eq: empirical marginal estimate} the marginal $p(y|w)$ is a weighted sum of sample contributions and so is the batch loss $\calL_w(\psi)$. Thus $\psi$ is straightforward to optimize by stochastic backpropagation, using unbiased mini-batch estimates $\nabla_\psi\tilde{\calL}_w(\psi)$.
 
\section{Experiments}
\label{sec: results}

\begin{figure}[t]
\includegraphics[width=\textwidth]{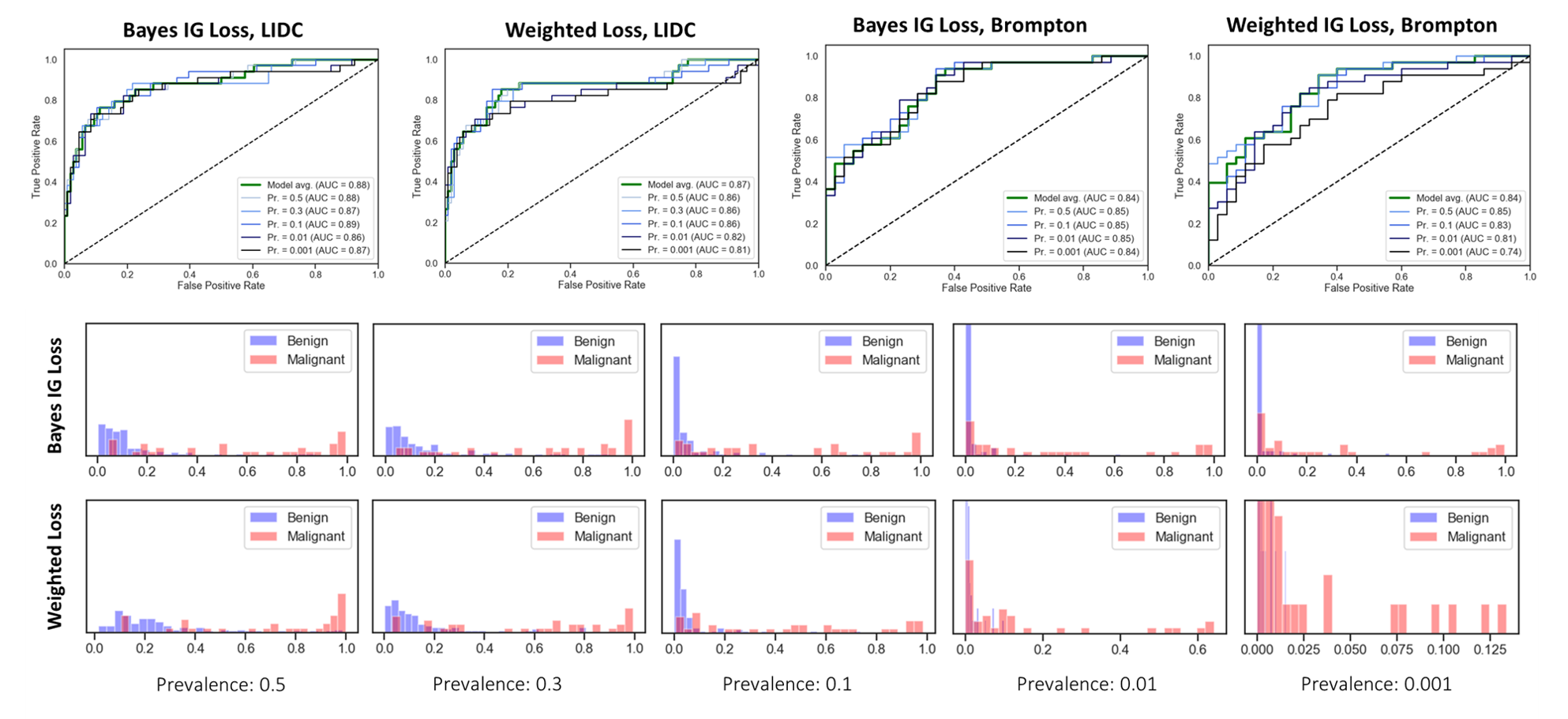}
\caption{Comparative view of the Bayesian bias corrected loss (Bayes IG) and the standard weighted loss. (Bottom two rows) Histogram (density) of predicted malignancy probability on hold-out dataset, for malignant (red) and benign (blue) samples. From left to right, the assumed true prevalence rate (passed to the losses at training time) is decreased. Note the change in probability scale for the weighted loss, which does not actually reflect the overlap of benign/malignant nodules. (Top row) ROC curves across the same prevalence ranges, on LIDC dataset and Brompton dataset. The Area Under the Curve degrades significantly at low prevalences for importance sampling.} \label{fig: roc LIDC}
\end{figure}

As a case study we consider the task of lung nodule malignancy assessment from CT images and/or available metadata (demographics, smoking). 
One interest of the proposed Bayesian loss is to account at training time for a mismatch between the apparent class distribution and the true prevalence. 
Importance sampling (a.k.a. weighted loss) serves as a natural benchmark.\\ 

\noindent
\textbf{Datasets.} We use the LIDC-IDRI dataset \cite{armato2011lung}, as well as an in-house dataset (Brompton). The LIDC-IDRI data includes $1000$ scans with one or more pinpointed nodules and corresponding annotations by multiple raters (typically $3$, $4$). The subjective malignancy score ranges from 1 (benign) to 5 (malignant), with $3$ indicating high uncertainty from the raters. 
The malignancy is predicted from $64^3$ patches extracted around the nodules. We experiment on $2$ variants of the dataset: (1) for binary classification, a dataset of $1407$ patches with a class imbalance of $1$ to $3$ in favor of benign nodules ($1065$ benign, $342$ malignant), for which nodules with an average rating of $3$ are excluded; and~(2) for subjective rating prediction, a dataset of $1086$ patches (marginal label distributions $\sim 0.075,0.2,0.45,0.2,0.075$) for which the raters' votes serve as a fuzzy ground truth. We considered several test time aggregation schemes w.r.t. raters for computation of confusion matrices (incl. majority voting or expected scores) with very similar trends across variants. The Brompton dataset consists of $679$ patches, with an equal balance of benign/malignant, and includes image-based (nodule diameter, solid/part-solid type, presence of emphysema) and non-imaging metadata (\eg age, sex, smoking status). \\

\noindent
\textbf{Architectures.} Experiments are reported on two variants of Deep Learning architectures as described in \cite{ciompi2017towards}. Triplets of orthogonal viewplanes (dimension $32\times 32$) are extracted at random from the $3$D patch, yielding a collection of $s$ views (here, $s=9$). Each $2$D view is passed through a singleview architecture to extract an $m$-dimensional (\eg, $m=256$) feature vector, with shared weights across views. The features are then pooled (min, max, avg elementwise) to derive an $m$-dimensional feature vector for the stack of views. A fully-connected layer outputs the logits that are fed to the likelihood (\eg, softmax) model. Because of the relatively small size of the datasets, we use low-level visual layers pretrained on vgg16 (we retain the two first conv+relu blocks of the pretrained model, and convert the first block to operate on grayscale images). The low-level visual module returns a $64$-channel output image for any input view, which is then fed to the main singleview model. In the first variant (ConvNet), the main singleview architecture consists in a series of $2$D strided convolutional layers (stride $2$, replacing the pooling layers in \cite{ciompi2017towards}), with ReLu activations and dropout ($p=.1$). The second variant replaces the convolutional layers with inception blocks. Despite variations in classifier performance, we have found similar trends to the ones reported here to hold across a range of architectures, from single-view fully convolutional classifiers to more complex gated models (\eg, using Gated Recurrent Units to encourage the model to implicitly segment the nodule). The models reported here were singled out as a trade-off between speed of experimentation for $k$ fold cross-validation and performance.\\

\noindent
\textbf{LIDC-IDRI.} The first experiment evaluates models trained for binary classification (benign/malignant) using either the proposed Bayesian loss or importance sampling. In both cases the minibatch is rebalanced (equal probability of benign/malignant samples). We set a ``true'' prevalence value for malignant nodules to either $0.3$ (very close to the actual dataset distribution) or $10^{-3}$ (which reflects a belief that there is a large amount of similar benign nodule for each case in the dataset). 
As per Table~\ref{table: LIDC - balanced minibatch}, the performance is similar for low class imbalance; but the behaviour significantly differs for higher imbalances. This is also noticeable from Fig~\ref{fig: roc LIDC}. The trends still hold true if the minibatches are sampled without rebalancing (from the batch distribution), cf. Table~\ref{table: LIDC - original batch statistics}.\\

\begin{table}[h!]
  \begin{center}
    \caption{Model performance on hold-out fold (repeated on $5$ folds out of $10$) when training using the proposed Bayesian bias-corrected loss (bayesIG) vs. importance sampling (wLoss). Accuracy is reported either as computed from the test sample (Acc.), weighted by prevalence (wAcc.) or balanced (BA). PPV/NPV: positive/negative predictive value. TPR/NPR: positive/negative rates.}
    \label{table: LIDC - balanced minibatch}
    \begin{tabular}{c@{\hskip .5em}|@{\hskip .5em}c@{\hskip .5em}|c|c|c|c|c|c|c|c}
      \multicolumn{2}{c|}{{}} &\textbf{exp.log-lik.} & \textbf{Acc.}& \textbf{wAcc.} & \textbf{BA} & \textbf{PPV}& \textbf{NPV}& \textbf{TPR}& \textbf{TNR}\\
      \toprule 
      \textbf{ConvNet}& \textbf{bayesIG} & $-0.3483$ & $0.88$ & $0.872$ & $0.826$& $0.843$& $0.885$& $0.712$& $0.941$\\ 
\textbf{prev. $= 0.3$}& \textbf{wLoss} & $-0.3488$& $0.87$ & $0.867$ & $0.831$& $0.807$& $0.893$& $0.740$& $0.921$\\ 
            	
      \midrule
      \textbf{InceptionNet}& \textbf{bayesIG} & $-0.39$ &$0.86$& $0.84$ & $0.78$& $0.79$& $0.86$& $0.64$& $0.92$\\ 
\textbf{prev. $= 0.3$}& \textbf{wLoss} & $-0.36$ & $0.87$ &$0.85$ & $0.81$& $0.81$& $0.87$& $0.69$& $0.93$\\ 
      \midrule
      \textbf{ConvNet}& \textbf{bayesIG} & $-0.027$ & $0.82$ &$0.9925$ & $0.66$& $0.67$& $0.999$& $0.33$& $0.99$\\ 
\textbf{prev. $= 10^{-3}$}& \textbf{wLoss} & $-0.0074$ & $0.74$ & $0.9989$ & $0.5$& 0 & $0.9989$& $0.$& $1.$\\ 
      \midrule
      \textbf{InceptionNet}& \textbf{bayesIG} & $-0.36$ & $0.83$ &$0.86$ & $0.81$& $0.81$& $0.93$& $0.69$& $0.928$\\ 
\textbf{prev.  $= 10^{-3}$}& \textbf{wLoss} & $-0.006$ & $0.74$ & $0.9989$ & $0.5$& $0$& $0.9989$& $0.$& $1.$\\ 
      \bottomrule 
    \end{tabular}
  \end{center}
\end{table}

\noindent
\textbf{Subjective rating prediction.} We use the same architectures and exchange the standard softmax likelihood for a likelihood model that better reflects the specificites of the rating. The model is plugged in Eq.~\eqref{eq: bias-corrected posterior} exactly as before. We draw inspiration from the ``stick-breaking'' likelihood~\cite{khan2012stick} to design an onion-peeling likelihood model, whose logic first assesses whether the nodule has clear benign (resp. malignant) characteristics (rating $1$ or $5$); if not, whether it has more subtle benign (malignant) characteristics ($2$ or $4$); and if not the nodule is deemed ambiguous (Appendix). For this experiment we use rebalancing at training time and assume the original dataset class probabilities to be the true prevalences. As expected for this task, the classifier accuracy is lower (Acc. $0.55$, BA. $0.45$). To account for the subjectivity of the rating, we also assess the off-by-one accuracy; which deems the prediction to be correct if the predicted label is off by no more than $1$ from the true label. Under this off-by-one scheme, Acc.: $0.94$, BA: $0.94$, true rates: $0.76$, $0.98$, $0.98$, $1.$, $0.93$. The resulting accuracy for benign/malignancy prediction is of $0.85$ (TPR: $0.92$, TNR: $0.76$, PPV: $0.80$, NPV: $0.90$).\\

\noindent
\textbf{Brompton dataset.} To make use of the available metadata, we couple the previous image-based architectures with a block that takes as input the non-imaging metadata. We use a two-layer predictor of the form $\sum_{u=1}^H K(g_u(x)) l_u(x)$, where $g_u, l_u$ are linear (affine) and $K$ is a radial basis kernel. The predictor outputs logits, which are aggregated with the image-based logits, then fed to a softmax likelihood for binary classification. Similar observations hold to LIDC-IDRI; whereby the Bayesian correction seems to be calibrated more consistently across a range of true prevalences compared to the weighted loss (Fig.~\ref{fig: roc LIDC}).

\section{Conclusion}
\label{sec: conclusion}

We introduced a family of loss functions to train models in the presence of sampling bias. The correction is derived following Bayesian principles and its explicit form draws connection with information gain. The case study points shows promising use cases for the approach. Beyond its natural integration in Bayesian Neural Networks, it seems well suited to handle problems with large class imbalance, as is common either for computer-aided diagnosis tasks or in \eg, segmentation. In future work we plan to investigate extensions of this approach to reweight samples adaptively based on current probability estimates.

\section*{Acknowledgements}

This research has received funding from the European Research Council (ERC) under the European Union's Horizon 2020 research and innovation programme (grant agreement No 757173, project MIRA, ERC-2017-STG). LL is funded through the EPSRC (EP/P023509/1).

%!TEX root = main.tex

\bibliographystyle{styles/bibtex/spmpsci}
\bibliography{bibliography}
\newpage

\appendix
\section{Technical appendix: proofs and derivations}
\label{sec: proofs}

\noindent
\textbf{Proof: $p(\cdot|x_\ast,X,Y)$ minimizes the BPR.} $\calL_{\text{Bayes}}$ immediately rewrites as 
\begin{align}
\calL_{\text{Bayes}}&=-\mathbb{E}_{p(y_\ast,x_\ast,X,Y,w)}[\log{q_{x_\ast,X,Y}(y_\ast)}]\, , \\
&=\text{KL}\left[\,p(\cdot|x_\ast,X,Y)\Vert q_{x_\ast,X,Y}(\cdot)\right] + \text{cst.}
\end{align}
The result follows from the properties of the Kullbach-Leibler divergence.\\

\noindent
\textbf{Proof of Eq.~\eqref{eq: bias-corrected posterior}.} The posterior can be expressed as the ratio $p(w|X,Y)=p(X,Y,w)/p(X,Y)$  of the joint probability and evidence. The latter is a constant of $w$. The tilde notation denotes distributions under the generative model of training data. Rewriting the joint distribution we get:
\begin{align}
p(w|X,Y) &\propto \tilde{p}(X|Y,w)\tilde{p}(Y|w)\tilde{p}(w) \, , \\
\, &\propto \tilde{p}(X|Y,w)\tilde{p}(Y)p(w) \, , \label{eq: Y ind w} \\
\, &\propto p(w)\cdot \textstyle \prod_n \tilde{p}(x_n|y_n,w) \, , \label{eq: iid}\\
\, &\propto p(w)\cdot \textstyle \prod_n p(x_n|y_n,w) \, , \label{eq: same conditional}\\
\, &\propto p(w) \cdot \prod_n \frac{p(y_n|x_n,w)p(x_n)}{p(y_n|w)} \, .
\end{align}
Eq.~\eqref{eq: Y ind w} uses $\tilde{p}(w)\eq p(w)$ and the independence $Y\!\perp\!\!\!\perp \!w$ in Fig.~\ref{fig: graphical models}(b). Eq.~\eqref{eq: iid} follows from the i.i.d. assumption $x_n{\perp\!\!\!\perp} x_{-n},y_{-n}\enspace|w$. The conditional $\tilde{p}(x_n|y_n,w)$ is unchanged in the label-based sampling, hence Eq.~\eqref{eq: same conditional}. The last line results from the application of Bayes' rule and the independence $x\!\perp\!\!\!\perp \!w$ in the true population's generative model. Eq.~\eqref{eq: bias-corrected posterior} ensues after dropping the constants of $w$. From the above we also see that for variational inference, the ELBO and its various usual expressions still hold.\\

\noindent
\textbf{Derivation of Eq.~\eqref{eq: empirical marginal estimate}.} Noting that $x \perp\!\!\!\perp w$ for the true population generative model, we get:
\begin{align}
p(y'|w) &= \textstyle \int_x p(y'|x,w) p(x) dx \, , \\
\, &= \textstyle \int_{x} p(y'|x,w) \left(\int_{y} p(x|w',y)p(y|w') dy\right) dx \, , \label{eq: x ind w'} \\
\, &= \int_{x} p(y'|x,w) \left(\int_{y} p(x|w_\ast,y)\tilde{p}(y)\frac{p(y|w_\ast)}{\tilde{p}(y)}dy \right) dx \, , \\
\, &= \int_{x} p(y'|x,w) \left(\int_{y} \frac{p_\calY(y)}{\tilde{p}(y)} \cdot p_{\calD'}(x,y)  dy\right) dx \, , \label{eq: rewriting} \\
\, &= \int_{x,y} \left(p(y'|x,w) \frac{p_\calY(y)}{\tilde{p}(y)}\right) \cdot p_{\calD'}(x,y)  d(x,y) \label{eq: final before empirical} \, , \\
\, &= \int_{y} p_\calY(y) \left(\int_x p(y'|x,w) \cdot p_{\calD'}(x|y) dx\right) dy \label{eq: final bis before empirical} \, .
\end{align} 
Eq.~\eqref{eq: x ind w'} holds for any value $w'$ by independence $x \perp\!\!\!\perp w$, and in particular for the true value $w_\ast$. Eq.~\eqref{eq: rewriting} uses $p_\calY(y)\!\triangleq\! p(y|w_\ast)$ and $p_{\calD'}(x,y)\!\triangleq\! p(x|w_\ast,y)\tilde{p}(y)$ from the generative model. Eq.~\eqref{eq: final before empirical} is turned into the empirical estimate of Eq.~\eqref{eq: empirical marginal estimate}. The notation $p_{\calD'}$ emphasizes the part that is approximated stochastically, using the minibatch sample distribution. Eq.~\eqref{eq: final bis before empirical} suggests an alternative minibatch estimate, whenever the minibatch contains at least $1$ sample from each class. Note that the distribution $\calD'$ of samples could optionally vary from minibatch to minibatch without affecting the validity of the derivations, and of the strategy outlined in the main text for the estimation of the marginal. 

\section{Additional tables}
\label{sec: tables}

Table~\ref{table: LIDC - original batch statistics} is the counterpart of Table~\ref{table: LIDC - balanced minibatch} when the minibatch samples are drawn i.i.d. following the batch statistics. Note that a prevalence of $.3$ for malignant nodules is very close to the apparent prevalence in the overall dataset. Therefore at this prevalence and in this setup without (minibatch) class rebalancing, the weighted loss behaves similarly to the standard (negative log-likelihood) loss. Since the dataset statistics are close to the assumed true statistics, this is a very favorable setting for the weighted/standard NLL loss. Note that the Bayesian IG loss behaves equally well.

\begin{table}[h!]
  \begin{center}
    \caption{Model performance on hold-out fold (average of $5$ folds out of $10$) when training using the proposed Bayesian bias-corrected loss (bayesIG) vs. importance sampling (wLoss). Accuracy is reported either as computed from the test sample (Acc.), weighted by prevalence (wAcc.) or balanced (BA). PPV/NPV: positive/negative predictive value. TPR/TNR: true positive/negative rates. In this variant, the minibatch sample distribution is that of the original batch (\textit{without} rebalancing) as opposed to equal distribution of all classes in the minibatches (\textit{with} rebalancing, with replacement). Since the dataset statistics are biased towards benign nodules, in this mode the weighted loss does not show as much degradation at low prevalence yet.}
    \label{table: LIDC - original batch statistics}
    \begin{tabular}{c@{\hskip .5em}|@{\hskip .5em}c@{\hskip .5em}|c|c|c|c|c|c|c|c|c}
      \multicolumn{2}{c|}{{}} &\textbf{log-lik.} & \textbf{Acc.}& \textbf{wAcc.} & \textbf{BA} & \textbf{PPV}& \textbf{NPV}& \textbf{TPR}& \textbf{TNR} & \textbf{AUC}\\
      \toprule 
      \textbf{ConvNet}& \textbf{bayesIG} & $-0.33$ & $0.89$ & $0.88$ & $0.83$& $0.86$& $0.89$& $0.72$& $0.95$& $0.91$\\ 
\textbf{prev. $= 0.3$}& \textbf{wLoss} & $-0.33$ & $0.89$ & $0.88$ & $0.83$& $0.84$& $0.89$& $0.72$& $0.94$& $0.90$\\ 
            	
      \midrule
      \textbf{InceptionNet}& \textbf{bayesIG} & $-0.36$ & $0.88$ & $0.85$ & $0.78$ & $0.86$& $0.85$& $0.61$& $0.96$& $0.90$\\ 
\textbf{prev. $= 0.3$}& \textbf{wLoss} & $-0.33$ & $0.88$ & $0.85$ & $0.79$& $0.84$& $0.86$& $0.64$& $0.95$ & $0.91$ \\ 
      \midrule
      \textbf{ConvNet}& \textbf{bayesIG} & $-0.008$ & $0.82$ &$0.999$ & $0.61$& $1.$& $0.999$& $0.2$& $1.$& $0.90$\\ 
\textbf{prev. $= 10^{-3}$}& \textbf{wLoss} & $-0.006$ & $0.77$ & $0.999$ & $0.50$& 0. & $0.999$& $0.$& $1.$& $0.86$\\ 
      \midrule
      \textbf{InceptionNet}& \textbf{bayesIG} & $-0.033$ & $0.82$ &$0.992$ & $0.65$& $0.67$& $0.999$& $0.31$& $0.99$ & $0.90$\\ 
\textbf{prev.  $= 10^{-3}$}& \textbf{wLoss} & $-0.006$ & $0.74$ & $0.999$ & $0.5$& $0.$& $0.999$& $0.$& $1.$ & $0.90$\\ 
      \bottomrule 
    \end{tabular}
  \end{center}
\end{table}

\end{document}